\documentclass[conference]{IEEEtran}
\IEEEoverridecommandlockouts
\usepackage{cite}
\usepackage{amsmath,amssymb,amsfonts}
\usepackage{algorithmic}
\usepackage{graphicx}
\usepackage[table,xcdraw]{xcolor}
\usepackage{textcomp}
\usepackage{xcolor}
\usepackage{multirow}
\def\BibTeX{{\rm B\kern-.05em{\sc i\kern-.025em b}\kern-.08em
    T\kern-.1667em\lower.7ex\hbox{E}\kern-.125emX}}
\begin{document}

\makeatletter
\newcommand{\linebreakand}{%
  \end{@IEEEauthorhalign}
  \hfill\mbox{}\par
  \mbox{}\hfill\begin{@IEEEauthorhalign}
}

\title{NORA: A Nephrology-Oriented Representation Learning Approach Towards Chronic Kidney Disease Classification}

\renewcommand{\footnoterule}{%
    \kern -3pt
    \hrule width 1.0\columnwidth
    \kern 2.6pt
}
\author{%
  Mohammad Abdul Hafeez Khan$^{1}$\thanks{Corresponding author: \texttt{mkhan@my.fit.edu}}%
  \quad Twisha Bhattacharyya$^{3}$%
  \quad Omar Khan$^{2}$%
  \quad Noorah Khan$^{2}$\\[0.1cm]
  \quad Alina Aziz Fatima Khan$^{4}$%
  \quad Mohammed Qutub Khan$^{5}$%
  \quad Sujoy Ghosh Hajra$^{1}$\\[0.2cm]
  $^{1}$Florida Institute of Technology, USA \quad
  $^{2}$University of California Riverside, USA\\[0.1cm]
  $^{3}$Edgewood Jr./Sr. High School, USA$ \quad
  ^{4}$Shadan Institute of Medicine Sciences, India\\[0.1cm]
  $^{5}$Riverside Nephrology Physicians, USA
}
 \maketitle

\begin{abstract}
Chronic Kidney Disease (CKD) affects millions of people worldwide, yet its early detection remains challenging, especially in outpatient settings where laboratory-based renal biomarkers are often unavailable. In this work, we investigate the predictive potential of routinely collected non-renal clinical variables for CKD classification, including sociodemographic factors, comorbid conditions, and urinalysis findings. We introduce the Nephrology-Oriented Representation leArning (NORA) approach, which combines supervised contrastive learning with a nonlinear Random Forest classifier. NORA first derives discriminative patient representations from tabular EHR data, which are then used for downstream CKD classification. We evaluated NORA on a clinic-based EHR dataset from Riverside Nephrology Physicians. Our results demonstrated that NORA improves class separability and overall classification performance, particularly enhancing the F1-score for early-stage CKD. Additionally, we assessed the generalizability of NORA on the UCI CKD dataset, demonstrating its effectiveness for CKD risk stratification across distinct patient cohorts.
\end{abstract}



\section{Introduction}

Chronic Kidney Disease (CKD) is a progressive condition characterized by the gradual loss of kidney function over time. Affecting an estimated 35.5 million individuals in the United States alone, CKD is often asymptomatic in its early stages and may go undetected until significant renal damage has occurred. This delayed diagnosis frequently leads to poor health outcomes, including cardiovascular disease, hospitalization, and premature mortality~\cite{b1}. Glomerular filtration rate (GFR) is regarded as the most reliable indicator of kidney function, signaling how effectively the kidneys filter waste from the blood. While GFR can be directly measured using inulin or other exogenous filtration markers, this approach is rarely used in clinical practice due to cost and complexity. As a result, clinical guidelines for CKD management rely heavily on the estimated GFR (eGFR), which is derived from serum creatinine levels, adjusted for age, sex, and race, and serves as a widely accepted indicator of renal function~\cite{b2}. Based on the eGFR value, patients are categorized into six stages of CKD. Stages~1--2 (eGFR~$\geq$~60) indicate normal to mildly reduced kidney function. Stage~3 is divided into~3a (eGFR 45--59) and~3b (30--44), reflecting moderate decline. Stage~4 (eGFR 15--29) indicates severe loss of kidney function, and Stage~5 (eGFR~$<$~15) corresponds to kidney failure. Finally, End-Stage Renal Disease (ESRD) refers to the phase where patients require long-term dialysis or post-transplant care.

While eGFR serves as the clinical standard for CKD staging, a range of routinely collected clinical variables from electronic health records (EHRs) also offer valuable predictive signals. Such information is particularly useful in outpatient or resource-limited settings where laboratory-based renal biomarkers may not be readily available. These variables are typically captured during patient intake, follow-up visits, or chronic disease management workflows.
They include sociodemographic factors (such as age, sex, and race), anthropometric measures (including height and weight), and chronic health conditions such as hypertension, diabetes mellitus, dyslipidemia, and diabetic nephropathy—among the most common contributors to kidney disease globally~\cite{b3,b4,b27,b28}. In addition, urinary abnormalities such as proteinuria and hematuria, when detected through routine urinalysis, can serve as early indicators of renal injury and are frequently incorporated into clinical assessments of kidney health~\cite{b5}. While these features do not directly measure renal filtration capacity, they are well-established risk factors for CKD and offer pathways for identifying individuals at elevated risk.

Prior works~\cite{b5,b6,b7,b8,b9,b10,b11,b12,b13,b14,b15,b16,b17,b18,b19} have applied machine learning (ML) methods to classify CKD, most notably using the widely studied UCI CKD dataset~\cite{b26}. While these models have achieved strong performance, they primarily rely on variables such as serum creatinine, blood urea, and hemoglobin—direct indicators of kidney function that may not always be accessible. In contrast, our work evaluates the predictive value of renal biomarker-agnostic clinical features for CKD classification. We leverage a clinic-based EHR dataset from Riverside Nephrology Physicians (RNP), which is \textit{a community-based, single-specialty practice in urban Southern California focused on the diagnosis and management of kidney diseases.} The RNP dataset includes de-identified patient records with CKD stage labels, along with sociodemographic, anthropometric, and routinely observed clinical variables. 

To establish a baseline for the RNP dataset, we first applied standard widely used ML algorithms. While these models achieved reasonable F1-score and accuracy, they struggled with two key challenges: the \textit{imbalanced distribution of the data} and the \textit{complex, nonlinear relationships among clinical variables.} In particular, linear and distance-based models failed to capture nuanced feature interactions, while even more flexible models like tree ensembles showed relatively low sensitivity to early-stage CKD. These limitations highlighted the need for a more expressive modeling approach.

\textbf{Contributions.} We propose the \textbf{N}ephrology-\textbf{O}riented \textbf{R}epresentation le\textbf{A}rning (\textbf{NORA}) approach, which applies supervised contrastive learning (SCL) with a nonlinear classifier to tabular nephrology data. NORA learns semantically meaningful representations from the clinical variables, improving CKD class separability in the latent space. To the best of our knowledge, this is the first application of SCL to tabular renal health records. We observe that the commonly used logistic regression (LR) with SCL, as proposed by Khosla et al.~\cite{b25}, underperforms in this setting due to complex feature interactions among patients. In contrast, using a Random Forest classifier significantly improves performance, especially on the early-stage CKD in the imbalanced RNP dataset—achieving a 10.1\% gain in F1-score compared to the SCL+LR baseline. Additionally, we evaluate NORA on the publicly available UCI CKD dataset~\cite{b26}, demonstrating its generalizability across distinct patient cohorts and clinical variable distributions.

\section{Related Works}

Traditional approaches to chronic kidney disease (CKD) risk prediction have relied on clinical equations and statistical models that incorporate laboratory-based biomarkers such as serum creatinine or estimated glomerular filtration rate (eGFR)~\cite{b5,b6,b7,b8,b9,b10}. While widely used, these methods have notable limitations. For instance, Dharmaratne et al.~\cite{b5} highlight the insensitivity of serum creatinine in early-stage detection, and Aoki et al.~\cite{b9} critique the Kidney Failure Risk Equation (KFRE) for its focus on late-stage outcomes. 

To address these limitations, machine learning (ML) has emerged as a powerful tool across domains~\cite{b19,b20}, including healthcare, where it has enabled advances in clinical decision support, and risk prediction~\cite{b21,b22,b23}. In the context of CKD, recent studies have applied ML techniques for disease classification and progression prediction using clinical and demographic features~\cite{b11,b12,b13,b14,b15,b16,b17,b18}. More specifically, these methods have been used to enhance early detection and risk stratification using EHR data~\cite{b7}, predict progressive kidney function decline through supervised learning~\cite{b8}, and capture complex, nonlinear feature interactions beyond traditional scoring systems~\cite{b9}. 

Mirza et al.~\cite{b11} and Pathak et al.~\cite{b15} reported strong results on the UCI CKD dataset~\cite{b26} using ensemble models such as Random Forest, Gradient Boosting, and CatBoost. Hossain et al.~\cite{b14} proposed a tri-phase ensemble strategy that combines bagging, boosting, and stacking, while Nikhila et al.~\cite{b18} incorporated additional evaluation metrics to improve diagnostic reliability. Other studies have proposed efficient CKD prediction pipelines to handle missing data, class imbalance, and feature selection, for reduced computational overhead~\cite{b12,b16}, and the integration of comorbid conditions to improve predictive performance~\cite{b17}. While these studies show promising results, most approaches rely on renal biomarkers for CKD classification and are often evaluated on a single dataset (e.g., UCI CKD), limiting broader clinical applicability.



\section{Dataset}

We use a de-identified dataset obtained from Riverside Nephrology Physicians (RNP). It consists of 960 patient records from 2014–2025, labeled with CKD stages (1–5) and End-Stage Renal Disease (ESRD), along with associated clinical variables. In consultation with nephrologists, we stratify the cohort into two clinically meaningful groups for modeling: patients with early-stage CKD (Stages 1–2; 183 samples) and those with moderate-to-advanced CKD (Stages 3-5 and ESRD; 777 samples). In this paper, we refer to CKD~$\leq$ 2 as \textbf{class 0} and CKD~$\geq$ 3 (including ESRD) as \textbf{class 1}. Table~\ref{tab:ckd_cohort_statistics} summarizes the dataset characteristics.

\begin{table}[htbp]
\centering
\scriptsize
\caption{Cohort Statistics for Riverside Nephrology Physicians (RNP) Dataset}
\begin{tabular}{|l|l|}
\hline
\textbf{Statistic} & \textbf{Value} \\
\hline
\multicolumn{2}{|c|}{\textbf{Dataset Overview}} \\
\hline
Total Samples & 960 \\
Class 1 (CKD $\geq$ 3) & 777 \\
Class 0 (CKD $\leq$ 2) & 183 \\
\hline
\multicolumn{2}{|c|}{\textbf{Age (years)}} \\
\hline
Mean $\pm$ Std & 70.0 $\pm$ 14.0 \\
Range & 23 -- 102 \\
25th / 50th / 75th Percentiles & 62 / 70 / 80 \\
\hline
\multicolumn{2}{|c|}{\textbf{Sex Distribution}} \\
\hline
Female & 483 (50.3\%) \\
Male   & 477 (49.7\%) \\
\hline
\multicolumn{2}{|c|}{\textbf{Race Distribution}} \\
\hline
Hispanic & 419 (43.6\%) \\
White & 357 (37.2\%) \\
Black or African American & 132 (13.8\%) \\
Asian & 52 (5.4\%) \\
\hline
\multicolumn{2}{|c|}{\textbf{Height (inches)}} \\
\hline
Available Samples & 908 \\
Mean $\pm$ Std & 65.2 $\pm$ 4.0 \\
Range & 49.5 -- 98.1 \\
\hline
\multicolumn{2}{|c|}{\textbf{Weight (lbs)}} \\
\hline
Available Samples & 908 \\
Mean $\pm$ Std & 182.4 $\pm$ 48.6 \\
Range & 64 -- 410 \\
\hline
\multicolumn{2}{|c|}{\textbf{Comorbidities}} \\
\hline
Hypertension & 711 (74.1\%) \\
Diabetes Mellitus & 355 (37.0\%) \\
Diabetic Nephropathy & 179 (18.6\%) \\
Proteinuria & 174 (18.1\%) \\
Hematuria & 8 (0.8\%) \\
Dyslipidemia & 90 (9.4\%) \\
\hline
\multicolumn{2}{|c|}{\textbf{CKD Stage Distribution}} \\
\hline
Stage 1 & 58 \\
Stage 2 & 125 \\
Stage 3 & 374 \\
Stage 4 & 226 \\
Stage 5 & 48 \\
End-Stage Renal Disease (ESRD) & 129 \\
\hline
\multicolumn{2}{|c|}{\textbf{Missing Values (\%)}} \\
\hline
Height & 5.4\% \\
Weight & 5.4\% \\
\hline
\end{tabular}
\label{tab:ckd_cohort_statistics}
\end{table}

The mean age of patients was 70.0 years, with a broad range spanning from 23 to 102 years. The sex distribution was nearly balanced, and the cohort was racially diverse, with 44\% Hispanic, 37\% White, 14\% Black or African American, and 5\% Asian individuals. Common comorbidities included hypertension (74.1\%), diabetes mellitus (37.0\%), and diabetic nephropathy (18.6\%), while urinary abnormalities like proteinuria and hematuria were present in 18.1\% and 0.8\% of patients, respectively. The dataset includes continuous variables such as age (in years), height (in inches), and weight (in pounds); nominal variables like sex (Female/Male) and race (Hispanic, White, Black or African American, Asian); and binary categorical variables for comorbidities including hypertension, diabetes mellitus, diabetic nephropathy, proteinuria, hematuria, and dyslipidemia, all recorded as Yes/No indicators.

Table~\ref{tab:ckd_stratified_statistics} further stratifies these statistics by classes 0 and 1. We used Mann--Whitney U tests for continuous variables due to non-normal distributions and Fisher’s exact tests for categorical variables. Patients in Class~0 were significantly younger than those in Class~1 ($p < 0.0001$), while no significant differences were observed in height or weight between the two classes ($p = 0.78$ and $p = 0.25$, respectively). Also, no significant difference in sex distribution was found between the groups ($p = 1.000$). Class~1 patients exhibited higher prevalence of hypertension ($p < 0.0001$), diabetic nephropathy ($p = 0.598$), and diabetes mellitus ($p = 0.006$), whereas proteinuria ($p = 0.0027$) was more common in Class~0. Hematuria ($p = 0.652$) and dyslipidemia ($p = 0.481$) were rare and not significantly different across classes. These comorbid conditions—particularly hypertension and diabetes mellitus—are well-established risk factors for CKD and strongly distinguish the two classes. Race-specific analysis revealed significant group differences, with Hispanic individuals more prevalent in Class~0 ($p < 0.0001$), and Asian individuals more represented in Class~1 ($p = 0.0018$). Black and White individuals were also slightly more represented in Class~1, though these differences were not statistically significant ($p = 0.056$ and $p = 0.175$, respectively).

\begin{table}[htbp]
\centering
\scriptsize
\caption{Cohort Statistics Stratified by CKD Status (Stage $\geq$ 3 vs. Stage $\leq$ 2)}
\begin{tabular}{|l|c|c|}
\hline
\textbf{Statistic} & \textbf{Class 0 (CKD $\leq$ 2)} & \textbf{Class 1 (CKD $\geq$ 3)} \\
\hline
\textbf{Sample Count} & 183 & 777 \\
\hline
\multicolumn{3}{|c|}{\textbf{Age (years)}} \\
\hline
Mean $\pm$ Std & 63.2 $\pm$ 14.4 & 71.6 $\pm$ 13.5 \\
\hline
\multicolumn{3}{|c|}{\textbf{Sex Distribution}} \\
\hline
Female (\%) & 92 (50.3\%) & 391 (50.3\%) \\
Male (\%)   & 91 (49.7\%) & 386 (49.7\%) \\
\hline
\multicolumn{3}{|c|}{\textbf{Race Distribution}} \\
\hline
Hispanic & 104 (56.8\%) & 315 (40.5\%) \\
White & 60 (32.8\%) & 297 (38.2\%) \\
Black or African American & 17 (9.3\%) & 115 (14.8\%) \\
Asian & 2 (1.1\%) & 50 (6.4\%) \\
\hline
\multicolumn{3}{|c|}{\textbf{Height (inches)}} \\
\hline
Mean $\pm$ Std & 65.2 $\pm$ 3.8 & 65.2 $\pm$ 4.0 \\
\hline
\multicolumn{3}{|c|}{\textbf{Weight (lbs)}} \\
\hline
Mean $\pm$ Std & 187.9 $\pm$ 53.6 & 181.1 $\pm$ 47.2 \\
\hline
\multicolumn{3}{|c|}{\textbf{Comorbidities (\%)}} \\
\hline
Hypertension & 109 (59.6\%) & 602 (77.5\%) \\
Diabetes Mellitus & 58 (31.7\%) & 297 (38.2\%) \\
Diabetic Nephropathy & 31 (16.9\%) & 148 (19.0\%) \\
Proteinuria & 48 (26.2\%) & 126 (16.2\%) \\
Hematuria & 2 (1.1\%) & 6 (0.8\%) \\
Dyslipidemia & 14 (7.7\%) & 76 (9.8\%) \\
\hline
\end{tabular}
\label{tab:ckd_stratified_statistics}
\end{table}

\section{Methodology}

\begin{figure*}[htbp]
\centerline{\includegraphics[width=\linewidth]{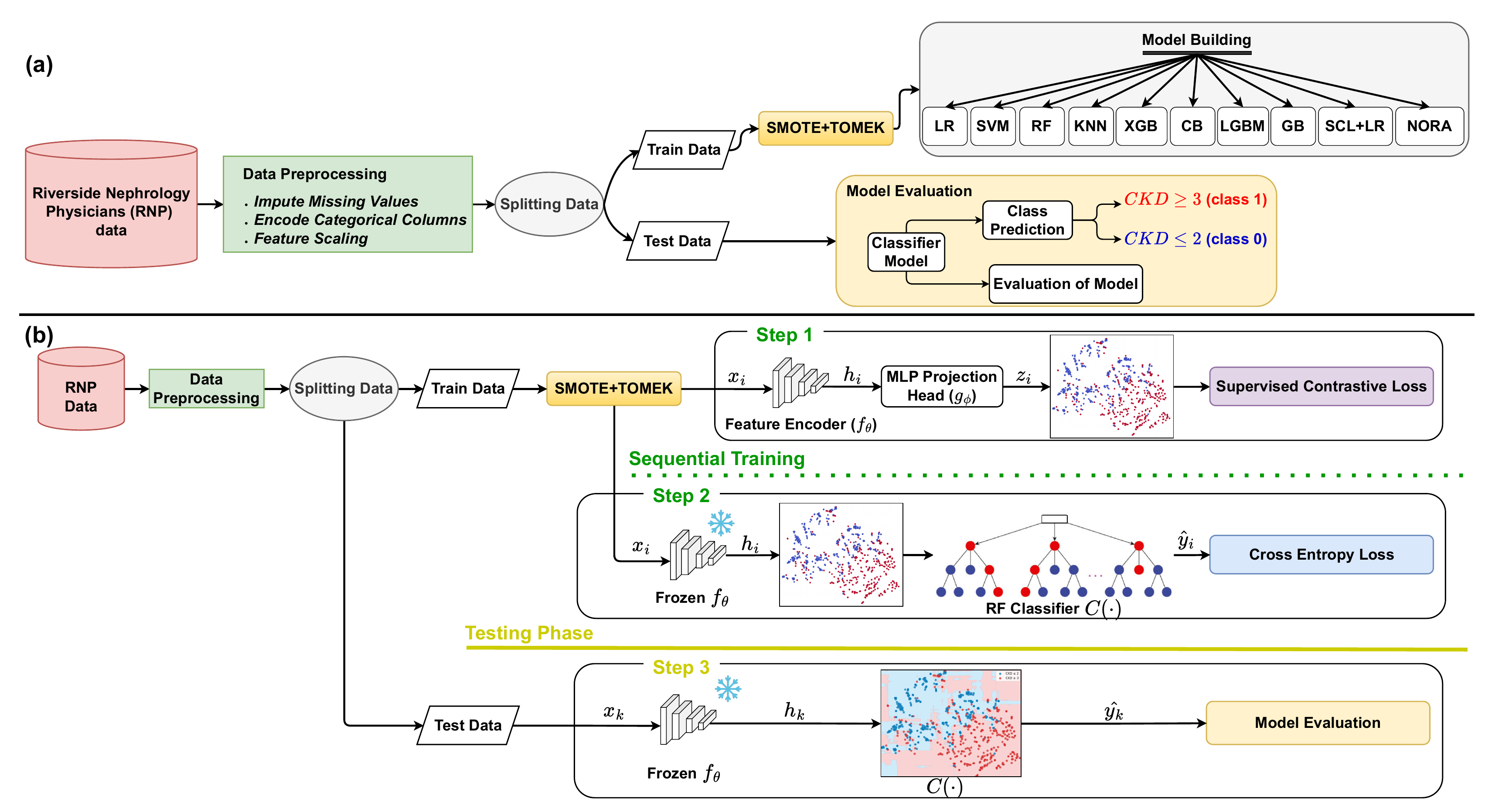}}
\caption{(a) End-to-end pipeline for benchmarking standard ML classifiers alongside our proposed method on tabular nephrology data. (b) Overview of our proposed \textbf{N}ephrology-\textbf{O}riented \textbf{R}epresentation le\textbf{A}rning (\textbf{NORA}) approach, which consists of two sequential training steps followed by the testing phase. \textbf{Step 1:} A supervised contrastive learning model is trained to encourage intra-class clustering and inter-class separation in the embedding space, grouping patients within each class (CKD Stage~$\leq$2 or CKD Stage~$\geq$3) while separating patients across the two classes.
\textbf{Step 2:} A Random Forest classifier is trained on the frozen encoder outputs. \textbf{Step 3:} The trained encoder and classifier are evaluated on a held-out test set.
}
\label{fig:nora_method}
\end{figure*}

Fig.~\ref{fig:nora_method} illustrates our methodology, consisting of (a) model benchmarking framework and (b) our proposed Nephrology-Oriented Representation leArning (NORA) approach.

\subsection{Model Benchmarking Framework}
\label{sec:base}
As shown in Fig.~\ref{fig:nora_method}(a), we begin by preprocessing the Riverside Nephrology Physicians (RNP) dataset. Missing values in numerical columns are filled using the mean of each feature. Categorical variables are handled as follows: sex and race are one-hot encoded to prevent artificial ordering, while binary comorbidities (hypertension, diabetes mellitus, etc.) are label-encoded as 0/1 values. All numeric features are standardized using z-score normalization via a standard scaler. Finally, the dataset is split into training and testing sets using stratified sampling to preserve the CKD class distribution.

\begin{figure}[htbp]
\centerline{\includegraphics[width=\linewidth]{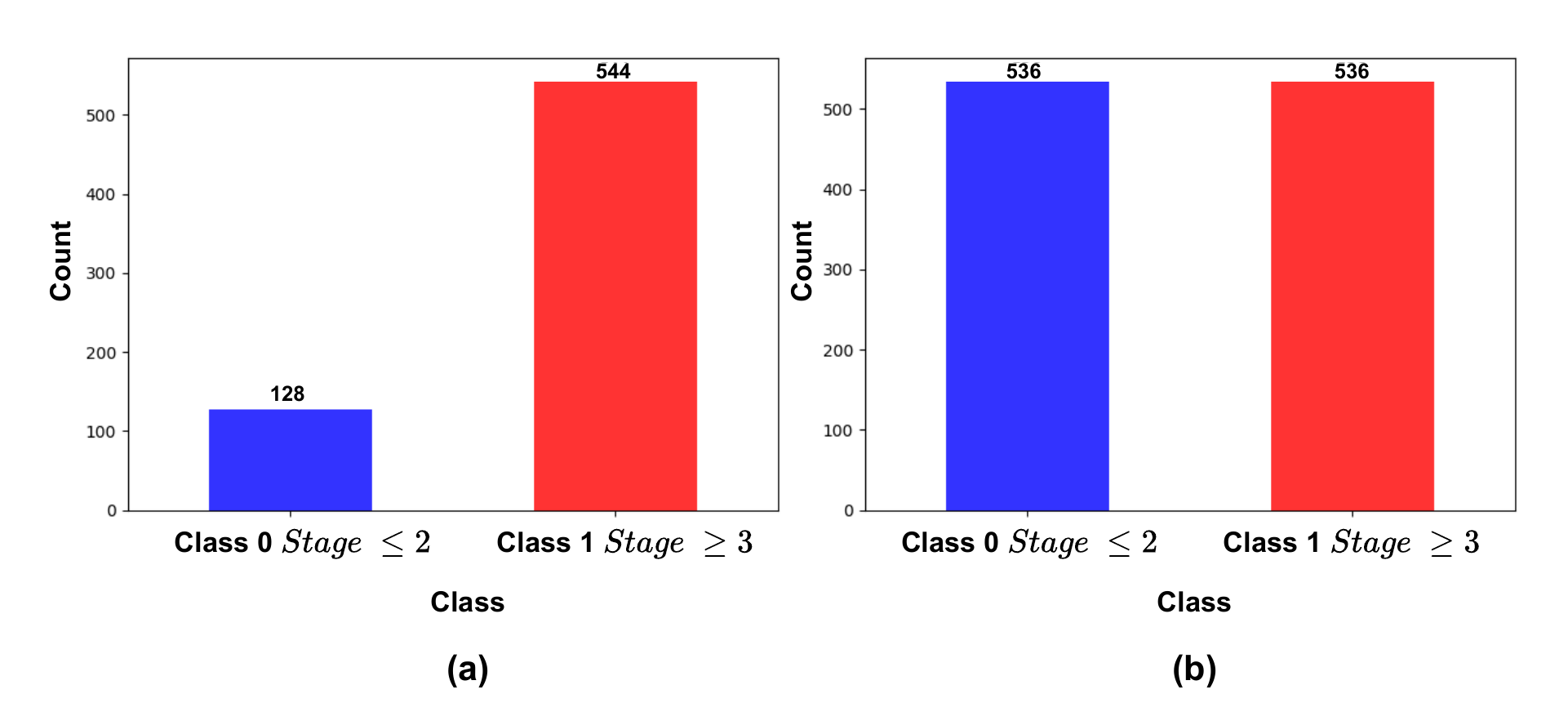}}
\caption{(a) Class distribution before SMOTE+Tomek (Train Set) (b) Class distribution after SMOTE+Tomek (Train Set). Blue bar denotes Class 0 and red bar denotes Class 1.}
\label{fig:smote_before}
\end{figure}


To address the significant class imbalance in our dataset, we apply the SMOTE+Tomek~\cite{b24} technique to the training set. From this point onward, we denote patients with CKD Stage~$\leq$~2 as \textbf{Class 0} and those with Stage~$\geq$~3 (including End-Stage Renal Disease (ESRD)) as \textbf{Class 1}. As illustrated in Fig.~\ref{fig:smote_before}(a), the training distribution is heavily skewed, with only 128 Class 0 samples compared to 544 Class 1 samples. This imbalance can lead to biased classifiers that favor the majority class. To mitigate this, we apply SMOTE to synthetically generate new samples for the minority class, and Tomek links to remove borderline noise from the majority class. After resampling, the two classes are balanced with 536 samples each (see Fig.~\ref{fig:smote_before}(b)), enabling improved generalization.

We begin by benchmarking a range of standard machine learning models on the preprocessed data, including Logistic Regression (LR), Support Vector Machine (SVM), Random Forest (RF), k-Nearest Neighbors (KNN), Gradient Boosting (GB), XGBoost (XGB), CatBoost (CB), and LightGBM (LGBM). Building on these results, we introduce our proposed method, NORA, in Section~\ref{nora}.

\subsection{Nephrology-Oriented Representation Learning Approach}
\label{nora}
Fig.~\ref{fig:nora_method}(b) illustrates our end-to-end pipeline for CKD classification using \textbf{N}ephrology-\textbf{O}riented \textbf{R}epresentation le\textbf{A}rning (\textbf{NORA}) approach. NORA leverages Supervised Contrastive Learning (SCL) to learn patient representations from tabular clinical data, followed by a nonlinear Random Forest (RF) classifier trained on the resulting embeddings.

Let the dataset be denoted as $\mathcal{D} = \{(\mathbf{x}_i, y_i)\}_{i=1}^{N}$, where each $\mathbf{x}_i \in \mathbb{R}^d$ is a clinical feature vector and $y_i \in \{0,1\}$ indicates class (CKD Stage~$\leq2$ vs.~Stage~$\geq3$ respectively). After preprocessing and SMOTE+Tomek resampling (see Sec.~\ref{sec:base}), we train the model as follows:

\paragraph{Representation Learning with SCL}
We use a multilayer perceptron (MLP) feature encoder $f_\theta: \mathbb{R}^x \rightarrow \mathbb{R}^h$ to transform input features into a latent feature space, followed by a projection head $g_\phi: \mathbb{R}^h \rightarrow \mathbb{R}^z$ that outputs $\ell_2$-normalized embeddings $\mathbf{z}_i = g_\phi(f_\theta(\mathbf{x}_i))$. These embeddings are optimized using the supervised contrastive loss, which brings together representations of patients from the same CKD class (intra-class) while pushing apart those from different class (inter-class). This objective leads to class-wise disentanglement in the embedding space, as shown in Fig.~\ref{fig:contrastive_embeddings}, where class 0 and class 1 patients become more distinctly clustered—enhancing downstream separability for classification. Formally, the supervised contrastive loss is defined as:
\begin{equation}
\mathcal{L}_{\text{SCL}} = \sum_{i=1}^{B} \frac{-1}{|P(i)|} \sum_{p \in P(i)} \log \frac{\exp(\mathbf{z}_i \cdot \mathbf{z}_p / \tau)}{\sum_{a \in A(i)} \exp(\mathbf{z}_i \cdot \mathbf{z}_a / \tau)}
\end{equation}
where $B$ is the batch size, $i$ denotes the anchor sample, $P(i)$ is the set of all positives for anchor $i$ (i.e., samples in the batch with the same label as $i$), and $A(i)$ is the set of all samples in the batch excluding $i$. The dot product $\mathbf{z}_i \cdot \mathbf{z}_p$ measures similarity, and $\tau$ is a temperature scaling parameter that controls the sharpness of the distribution.

\paragraph{Classification with Random Forest}
After SCL, we remove the projection head and freeze the encoder $f_\theta$. Each patient is then embedded into the latent feature space $\mathbf{h}_i = f_\theta(\mathbf{x}_i)$, which serves as input to a Random Forest (RF) classifier $C(\cdot)$ for binary CKD classification. RF~\cite{b29} is an ensemble of decision trees that partitions the input space using axis-aligned splits and aggregates the outputs of multiple trees to produce robust, nonlinear decision boundaries. 

We initially evaluated a logistic regression (LR) classifier on top of the learned representations, following Khosla et al.~\cite{b25} direction, but its performance was limited. Although the embedding space shows improved class separation after SCL, the clusters remain nonlinearly separable due to complex and overlapping feature distributions in the RNP data. While SMOTE+Tomek helps mitigate class imbalance, the synthetically generated minority samples may not fully capture the clinical heterogeneity of early-stage CKD (class 0). In contrast, RF better models these complex relationships in the latent space, as illustrated in Fig.~\ref{fig:classifier_boundaries}. The classifier is trained using binary cross-entropy loss:
\begin{equation}
\hat{y}_i = C(\mathbf{h}_i), \quad 
\mathcal{L}_{\text{CE}} = - \sum_{i=1}^{N} \left[ y_i \log \hat{y}_i + (1 - y_i) \log (1 - \hat{y}_i) \right]
\end{equation}


\paragraph{Inference Phase}
At test time, an unseen input $\mathbf{x}_k$ is passed through the frozen encoder $f_\theta$ to obtain its latent representation $\mathbf{h}_k = f_\theta(\mathbf{x}_k)$, which is then fed into the trained classifier to produce the final prediction: $\hat{y}_k = C(\mathbf{h}_k)$.


\begin{figure}[htbp]
\centerline{\includegraphics[width=1.1\linewidth]{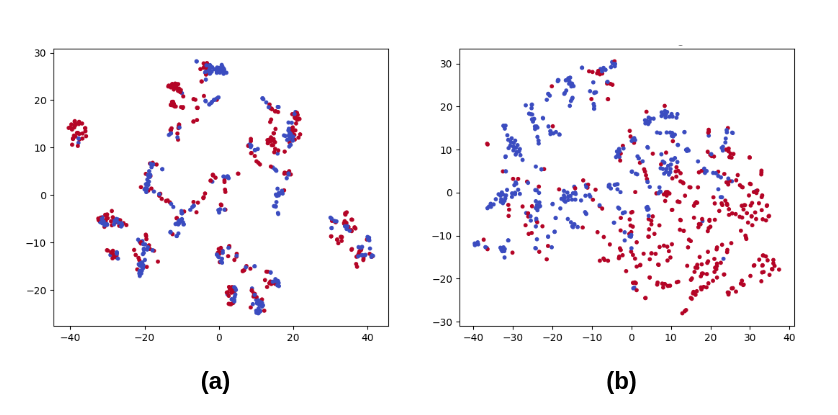}}
\caption{t-SNE visualizations. (a) Train embeddings before contrastive learning. (b) Train embeddings after contrastive learning, showing intra-class clustering and clearer inter-class separation. Blue points denote Class 0 (CKD Stage~$\leq$~2) and red points denote Class 1 (CKD Stage~$\geq$~3).
}
\label{fig:contrastive_embeddings}
\end{figure}

\begin{figure}[htbp]
\centerline{\includegraphics[width=\linewidth]{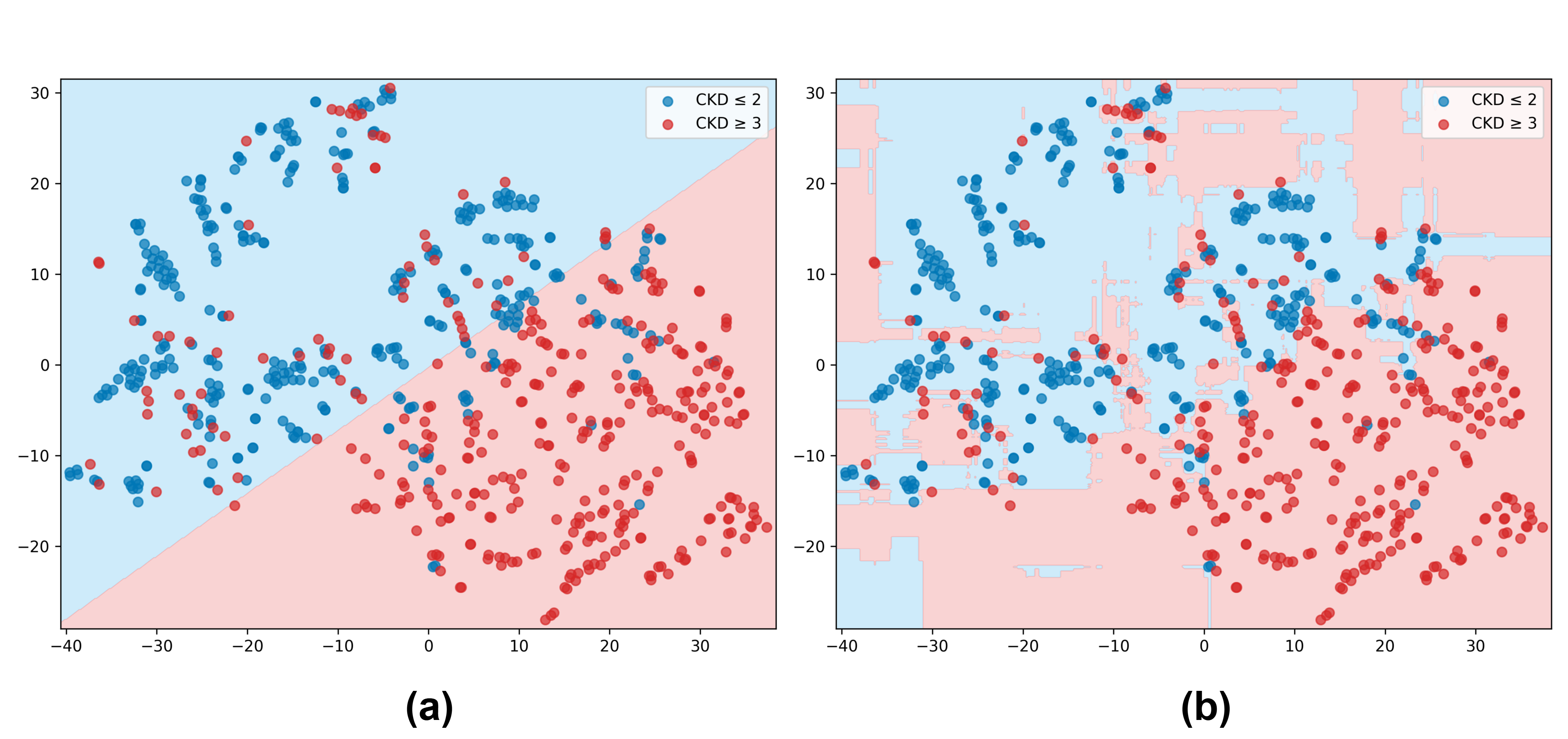}}
\caption{(a) Decision boundary learned by logistic regression. (b) Decision boundary learned by Random Forest, capturing more complex class separation.}
\label{fig:classifier_boundaries}
\end{figure}

\section{Experimental Results}
\label{sec:exp}

\begin{table*}[htbp]
\centering
\caption{Per-Class, Macro, Weighted Averages, and Accuracy for CKD Classification (Class 0 :(CKD~$\leq$ 2), Class 1: CKD~$\geq$ 3)}
\begin{tabular}{|l|ccc|ccc|ccc|ccc|c|}
\hline
\multirow{2}{*}{\textbf{Model}} & \multicolumn{3}{c|}{\textbf{Class 0}} & \multicolumn{3}{c|}{\textbf{Class 1}} & \multicolumn{3}{c|}{\textbf{Macro Avg}} & \multicolumn{3}{c|}{\textbf{Weighted Avg}} & \textbf{Accuracy} \\
\cline{2-14}
 & Prec. & Rec. & F1 & Prec. & Rec. & F1 & Prec. & Rec. & F1 & Prec. & Rec. & F1 & \\
\hline
Logistic Regression (LR) & 58.0 & 33.0 & 42.0 & 71.0 & 87.0 & 78.0 & 64.0 & 60.0 & 60.0 & 66.5 & 68.5 & 65.8 & 68.5 \\
Support Vector Machine (SVM)                 & 63.0 & 35.0 & 45.0 & 72.0 & 89.0 & 80.0 & 68.0 & 62.0 & 62.0 & 69.1 & 70.5 & 67.8 & 70.5 \\
K-Nearest Neighbors (KNN)                 & 66.0 & 56.0 & 61.0 & 79.0 & 85.0 & 82.0 & 72.0 & 71.0 & 71.0 & 74.4 & 75.0 & 74.5 & 75.0 \\
Random Forest (RF)      & 79.0 & 63.0 & 70.0 & 82.0 & 91.0 & 87.0 & 81.0 & 77.0 & 78.0 & 81.4 & 81.6 & 81.0 & 81.6 \\
Gradient Boosting (GB)   & 71.0 & 53.0 & 60.0 & 78.0 & 88.0 & 83.0 & 74.0 & 71.0 & 72.0 & 75.5 & 76.2 & 75.2 & 76.2 \\
XGBoost (XGB)             & 79.0 & 70.0 & 74.0 & 85.0 & 90.0 & 88.0 & 82.0 & 80.0 & 81.0 & 83.2 & 83.4 & 83.2 & 83.4 \\
LightGBM            & 79.0 & 70.0 & 74.0 & 85.0 & 90.0 & 88.0 & 82.0 & 80.0 & 81.0 & 83.0 & 83.3 & 83.0 & 83.0 \\
CatBoost (CB)            & 78.0 & 62.0 & 69.0 & 82.0 & 91.0 & 86.0 & 80.0 & 77.0 & 78.0 & 80.7 & 81.0 & 80.4 & 81.0 \\
\rowcolor{gray!10} SCL+LR & 68.0 & 70.0 & 69.0 & 84.0 & 83.0 & 84.0 & 76.0 & 77.0 & 76.0 & 79.0 & 79.0 & 79.0 & 79.0 \\
\rowcolor{green!10} \textbf{NORA}       & \textbf{81.0} & \textbf{72.0} & \textbf{76.0} & \textbf{86.0} & \textbf{91.0} & \textbf{88.0} & \textbf{83.0} & \textbf{82.0} & \textbf{82.0} & \textbf{84.2} & \textbf{84.4} & \textbf{84.2} & \textbf{84.4} \\
\hline
\end{tabular}
\label{tab:ckd_full_metrics_with_acc}
\end{table*}

We primarily evaluate all models on the Riverside Nephrology Physicians (RNP) dataset and additionally assess the generalization performance of our proposed method on the publicly available UCI CKD dataset~\cite{b26}. All models are evaluated using five-fold stratified cross-validation to ensure robustness, and the reported metrics represent the average performance across folds.

\paragraph{Model Architecture}
The feature encoder in our \textbf{N}ephrology-\textbf{O}riented \textbf{R}epresentation le\textbf{A}rning (\textbf{NORA}) approach is implemented as a 3-layer multilayer perceptron (MLP) with hidden dimensions \([128, 64, 32]\), each followed by ReLU activation. The projection head is a 2-layer MLP that maps the encoder output to a 32-dimensional embedding. It is followed by $\ell_2$ normalization to produce unit-length vectors, as required for supervised contrastive learning (SCL). The encoder and projection head are jointly trained using the Adam optimizer with a learning rate of $1 \times 10^{-3}$, batch size of 64, and trained for 100 epochs. The temperature parameter $\tau$ for SCL is set to 0.07. After training, the projection head is removed, and a downstream Random Forest (RF) classifier is trained on the frozen encoder embeddings. The RF classifier uses 100 trees (\texttt{n\_estimators=100}) and a maximum depth of 10 (\texttt{max\_depth=10}).

\paragraph{Benchmarking}
We evaluate the performance of NORA against SCL representations followed by a logistic regression (LR) classifier (SCL+LR), and the standard machine learning models including LR, support vector machine (SVM), Random Forest (RF), XGBoost (XGB), k-nearest neighbors (KNN), CatBoost (CB), Gradient Boosting (GB), and LightGBM. All models are trained using default hyperparameters unless otherwise specified. 

\paragraph{Evaluation Metrics}
We report Accuracy, Precision, Recall, and F1-Score to assess classification performance.

\subsection{Quantitative Results}

\paragraph{RNP Dataset}

Table~\ref{tab:ckd_full_metrics_with_acc} presents the classification performance of the models for predicting class 0 (CKD~$\leq$ 2)  vs. class 1 (CKD~$\geq$ 3) on the RNP dataset. 

Among the baseline models in Table~\ref{tab:ckd_full_metrics_with_acc}, linear and distance-based classifiers exhibit significant performance gaps across the two classes. LR and SVM attain low F1-scores for class 0 (42.0 and 45.0, respectively), primarily due to poor recall (33.0\% and 35.0\%), indicating limited ability to detect early-stage CKD. This underperformance is due to their reliance on linear decision boundaries, which fail to capture the complex feature interactions inherent in the RNP data. KNN performs moderately (class 0 F1: 61.0, class 1 F1: 82.0) but lacks the capacity to model latent structure or adaptively weight features, leading to suboptimal separation in high-dimensional space. In contrast, RF achieves more balanced performance (class 0 F1: 70.0, class 1 F1: 87.0), reflecting its robustness to feature heterogeneity and ability to partition the feature space nonlinearly.

Among boosting-based models, XGB and LightGBM achieve higher F1-scores for class 0 (74.0) compared to LR, SVM, and RF, however, they still fall short of the performance achieved by NORA. Both models match NORA in class 1 performance (\textbf{88.0}), indicating their effectiveness in capturing patterns associated with advanced CKD, while remaining comparatively less sensitive to early-stage cases. CB shows a further drop in class 0 F1-score (69.0) despite strong precision for class 1, indicating that enhanced categorical handling alone does not compensate for weak early-stage signal representation. GB performs worse, with class 0 and class 1 F1-scores of 60.0 and 83.0, respectively, likely due to less effective regularization and suboptimal tree growth compared to XGB and LightGBM. These results highlight that while boosting approaches generally benefit from their ensemble structure and ability to model nonlinearities, they may require more class-balanced and diverse data to improve early CKD (class 0) detection. While SMOTE+Tomek helps address imbalance, the synthetic minority samples may not fully capture early-stage heterogeneity, a limitation more effectively handled by our NORA approach.

NORA achieves the strongest class-wise performance, with F1-scores of \textbf{76.0} for class~0 and \textbf{88.0} for class~1, indicating balanced detection of both early-stage and advanced CKD. Replacing the RF with a logistic regression (LR) classifier (SCL+LR) results in lower scores (class~0 F1: 69.0, class~1 F1: 84.0), reflecting a 10.1\% overall drop in F1-score and highlighting the benefit of using a nonlinear ensemble model to capture better decision boundaries in the contrastively learned embedding space (see Fig.~\ref{fig:classifier_boundaries}). As a tree-based method, RF can better capture nonlinear interactions and hierarchical feature splits, making it more effective at leveraging the structure of the learned representations where class boundaries are not linearly separable. In Fig.~\ref{fig}, we present the ROC curves and observe that NORA achieves an AUC of \textbf{0.888}, slightly higher than LightGBM (0.886) and XGB (0.885), while outperforming all other baselines.
Additionally, in Table~\ref{tab:ckd_full_metrics_with_acc}, weighted averages represent overall performance by accounting for class imbalance in the test data, while macro averages treat both classes equally. NORA performs well under both metrics, and also achieves the highest accuracy of \textbf{84.4\%}.


\begin{figure}[htbp]
\centerline{\includegraphics[width=\linewidth]{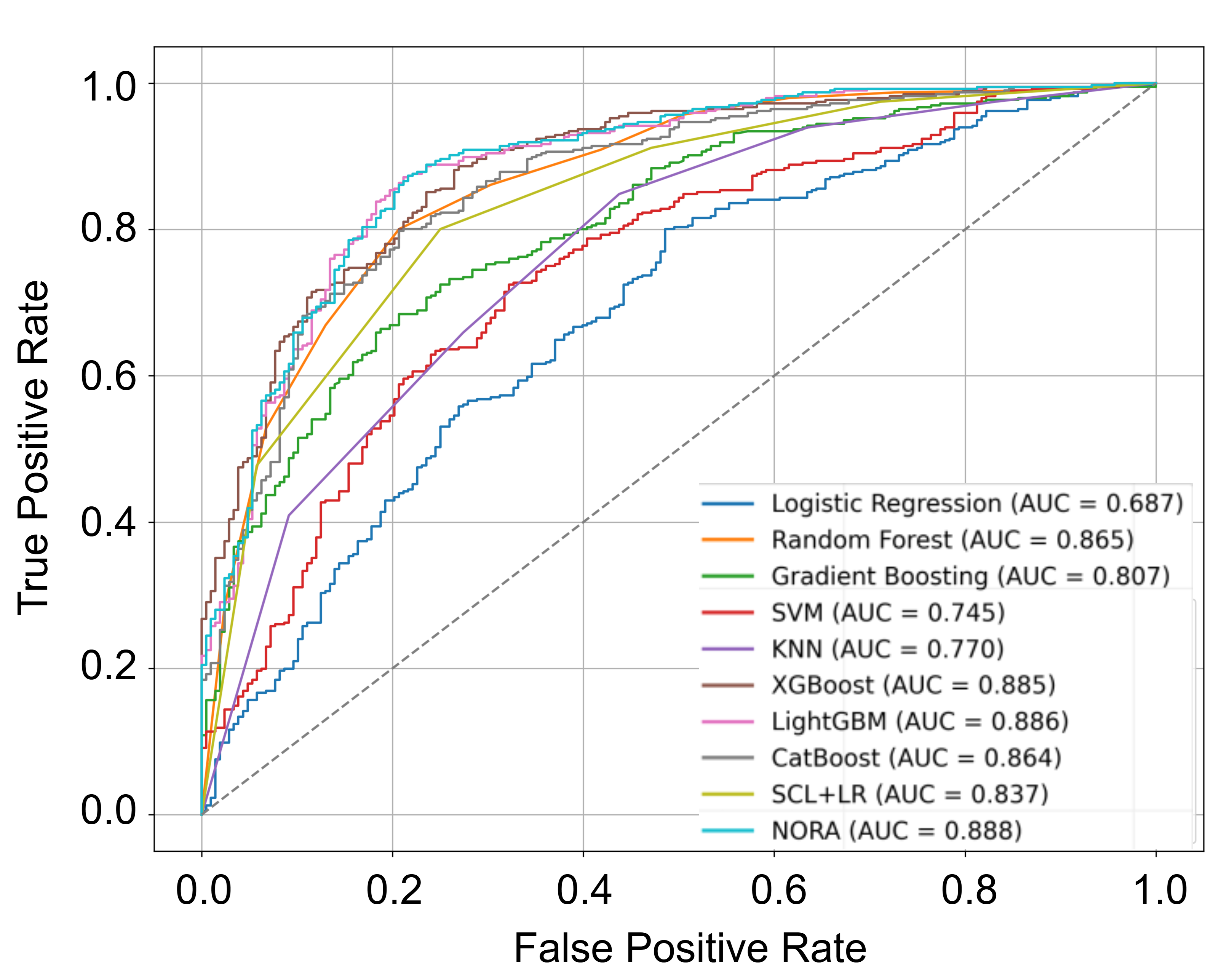}}
\caption{ROC curve comparison for CKD classification on the RNP dataset. The legend indicates each model's name, line color, and corresponding AUC.}
\label{fig}
\end{figure}

\begin{table}[htbp]
\caption{Performance of baseline models and NORA on the UCI CKD dataset. NORA achieves competitive results in both F1-score and accuracy.}
\begin{center}
\begin{tabular}{|l|c|c|c|c|}
\hline
\textbf{Model} & \textbf{Precision} & \textbf{Recall} & \textbf{F1-Score} & \textbf{Accuracy} \\
\hline
Logistic Regression & 96.9 & 97.0 & 97.0 & 96.9 \\
SVM & \textbf{100} & 96.9 & 98.4 & \textbf{98.0} \\
KNN & \textbf{100} & 97.0 & \textbf{98.5} & 96.5 \\
Random Forest & 98.2 & 96.9 & 97.5 & 97.0 \\
Gradient Boosting & 95.5 & \textbf{98.5} & 96.9 & 96.2 \\
XGBoost & 97.0 & \textbf{98.5} & 97.8 & 97.6 \\
LightGBM & 98.5 & \textbf{98.5} & \textbf{98.5} & \textbf{98.0} \\
CatBoost & 98.4 & 96.9 & 97.7 & 97.0 \\
\rowcolor{gray!10} SCL+LR & 97.5 & 97.8 & 97.6 & 97.6 \\
\cellcolor{green!10}\textbf{NORA} & 98.6 & \cellcolor{green!10}\textbf{98.5} & \cellcolor{green!10}\textbf{98.5} & \cellcolor{green!10}\textbf{98.0}\\
\hline
\end{tabular}
\label{tab:ckd_class1_only}
\end{center}
\end{table}

\paragraph{UCI CKD Dataset~\cite{b26}}  
This dataset consists of 23 clinical attributes spanning numerical and categorical types, along with a binary CKD label (yes/no). We apply the same pipeline described in Fig.~\ref{fig:nora_method}(a), substituting the input with the UCI CKD dataset and applying dataset-specific preprocessing steps. As shown in Table~\ref{tab:ckd_class1_only}, nearly all models achieve high performance, with F1-scores exceeding 96\%. This can be attributed to the low class imbalance (250 CKD vs. 150 non-CKD cases out of 400) and the presence of renal biomarker features such as serum creatinine, blood urea, and hemoglobin in the dataset. We observe that NORA performs competitively in this setting as well, matching strong baselines like LightGBM, SVM, and KNN, and achieving an F1-score and accuracy of \textbf{98.5}\% and \textbf{98.0}\%, respectively. 


\section{Limitations}

Our study has several limitations. First, to address class imbalance in the RNP dataset, we utilized SMOTE+Tomek to synthetically balance the training data, which improved performance but is less effective in sparsely populated regions of the feature space. Second, in consultation with nephrologists, we stratified the cohort into two classes: early-stage CKD (Stages 1–2) and moderate-to-advanced CKD (Stages 3 and above); however, stage-wise classification remains a key clinical goal, which will require larger and more balanced datasets. Third, the RNP dataset includes only adult patients aged 18 and older. Consequently, the trained models are not applicable to pediatric populations, where CKD may present and progress differently. In the future, incorporating pediatric data is necessary to enable broader model applicability. Fourth, conventional ML approaches exhibit limited generalizability in real-world settings with scarce data availability. In future work, we may explore few-shot and self-supervised learning approaches~\cite{b23} to enhance representation learning and improve generalization under limited data conditions.

\section{Conclusion}

We conducted our study on the Riverside Nephrology Physicians (RNP) dataset, which includes non-renal clinical variables and CKD staging for a diverse patient population. To leverage this tabular EHR data for CKD classification, we proposed the \textbf{N}ephrology-\textbf{O}riented \textbf{R}epresentation le\textbf{A}rning (\textbf{NORA}) approach to learn discriminative patient representations for improved classification. NORA outperforms baseline models in F1-score and accuracy, notably improving performance on early-stage CKD cases. It also generalizes well to the UCI CKD dataset, demonstrating its robustness across diverse patient cohorts and clinical feature distributions.

\section{Ethical Considerations}

This research used de-identified clinical data from Riverside Nephrology Physicians and received approval from the Institutional Review Board at the Florida Institute of Technology.

\end{document}